# Stochastic Streets: A Walk Through Random LLM Address Generation in four European Cities

Tairan Fu[1], David Campo-Nazareno[2,3], Javier Coronado-Blázquez[4], Javier Conde[2,5] Pedro Reviriego[2,5], and Fabrizio Lombardi[6]

[1]*Politecnico di Milano, Italy*

[2]*Universidad Politécnica de Madrid, Spain*

[3]*FIWARE foundation, Berlin, Germany*

[4]*Telefónica Tech, Madrid, Spain*

[5]*Information Processing and Telecommunications Center (IPTC), Spain*

[6]*Northeastern University, Boston, USA*

Abstract: Large Language Models (LLMs) are capable of solving complex math problems or answer difficult questions on almost any topic, but can they generate random street addresses for European cities?

Large Language Models (LLMs) have shown impressive performance across a wide range of tasks, such as answering questions on virtually any topic. However, there remain areas in which their performance falls short, for example, seemingly simple tasks like counting the letters in a word. In this column, we explore another such challenge: generating random street addresses for four major European cities. Our results reveal that LLMs exhibit strong biases, repeatedly selecting a limited set of streets and, for some models, even specific street numbers. Surprisingly, some of the more prominent and iconic streets are not selected by the models and the most frequent numbers in the responses lack any clear significance. This raises intriguing questions about how LLMs represent and retrieve geographic or structured information. One might expect these models to favor well-known streets, yet their outputs suggest that this is not always the case. Overall, random address generation emerges as another task in which LLMs still have notable room for improvement.

**LLMs Struggle with Simple Tasks**

Despite achieving unprecedented performance in many areas and being able to answer difficult questions on almost any topic [1], there are simple tasks that LLMs are unable to complete. A well-known example is counting letters in a word, many models cannot count the number of 'r' in strawberry [2] and most models also struggle to perform arithmetic calculations [3]. The failure in counting letters can, to some extent, be explained by the fact that LLMs operate on tokens, sequences of characters, rather than on individual letters. Consequently, they do not *perceive* letters in isolation within the input text, making direct letter counting challenging. LLMs generate text recursively by predicting the next token based on preceding tokens. In the case of numbers, it means that they "learn" specific operations, so "123 + 407 = 430" is learned independently from "12 + 31 = 43" as the model learns to predict the next token based on previous ones, but it does not have the "concept" addition. This can also partly explain why they struggle to perform arithmetic operations with arbitrary numbers, as those operations are not commonly found in the original texts used to train the models.

Another task which LLMs find hard is to generate random numbers. This was studied in [4] that found that although LLMs are fundamentally probabilistic models, when prompted to generate random numbers, they often produce deterministic or predictable patterns rather than truly random sequences. This may occur because their output is conditioned on learned statistical patterns from training data rather than on intrinsic randomness. The randomness quality depends heavily on the prompt wording and the model variant used. Slight changes in prompt phrasing can lead to different degrees of randomness, indicating that LLMs are sensitive to context cues rather than generating numbers purely by chance. As discussed, LLMs do not have built-in mechanisms for generating true random numbers. Instead, they predict the most probable next token given the prompt, which limits their ability to produce outputs that follow a truly stochastic distribution.

**LLM Random Addresses Generation**

A further twist is to test LLMs when generating random samples of data that are linked to the physical world and have some structure, a simple example is addresses composed of the street name and number. Generating random addresses can be useful in a wide range of applications, from logistics or public safety to tourism or research. For example, it is useful to simulate routes and optimize traffic management, but also to plan and evaluate emergency response scenarios for public services, such as ambulances, fire-fighters or police. Address generation is also useful for some tourism applications, for example for random street tourism, or when LLMs are used to plan routes for tourism in cities because users increasingly rely on LLMs for advice and guidance. In research, address generation is valuable to validate address matching algorithms and geocoding schemes, or for synthetic data generation, or data augmentation of datasets of addresses.

For addresses, the generation process has to rely on the internal model knowledge[1] of the streets for a given city, making the task harder. For the numbers, we could see if the biases observed in [4] are also present in the results, or if the selection of the numbers follows different trends. The first step is to select a group of cities for the study; we selected four important European cities: Amsterdam, Madrid, Paris and Rome. These cities are in different countries, each having a different language, size, population and surface, and are all major tourist destinations[2], so they should be well represented in the training data of the LLMs. Moreover, they provide a small set on which to test if LLM performance varies across cities and languages while making the task easier for LLMs compared to less well-known cities.

---

[1] In the experiments, the LLMs answer without recourse to external tools or help, just based on their internal knowledge.

[2] The four cities are in the top 10 destinations according to Euromonitor: https://www.euromonitor.com/press/press-releases/december-2024/euromonitor-international-reveals-worlds-top-100-city-destinations-for-2024

As with the cities, a selection must be made on the models evaluated to keep the study manageable. The models selected are summarized in Table 1, they include models from five different companies, both proprietary and open-weight, and are intended to represent the state of-the-art models available at the time of writing this column both proprietary (GPT-4.1, Gemini 2.5 Flash) and open-weight that can be run locally on a Graphics Processing Unit (GPU).

*Table 1. Models selected for the Evaluation*

| Model | Developer | License Type | Description |
| --- | --- | --- | --- |
| GPT-4.1 | OpenAI | Proprietary | An advanced model with strong reasoning and coding skills, improving on GPT-4's performance. |
| Gemini 2.5 Flash | Google | Proprietary | A fast, efficient variant of Gemini for quick, low-latency tasks. |
| Gemma 3-12B | Google | Open-weights | An open-weight model from the Gemini family, tuned for research and fine-tuning. |
| LLaMA 3.1–8B | Meta AI | Open-weights (with restrictions) | A refined version of LLaMA 3 open-weight models focused on instruction-following and alignment. |
| Mistral-7B | Mistral AI | Open-weights | A compact, high-performance, open-weight model with strong multilingual and reasoning abilities. |
| Qwen3-8B | Alibaba Cloud | Open-weights (with restrictions) | An open-weight model optimized for English and Chinese, strong in dialogue and code tasks. |

To generate the addresses, we must select the model configuration and prompt. The default parameters are used for all models except for the temperature that was set to one in all cases. This allows the model to sample on a uniform token probability distribution while controlling hallucinations or incoherent generation that are common for higher temperature values. These settings are intended to make results across models comparable, and to reflect the configurations utilized by users. As for the prompt, the analysis in [4] has shown that it can have a significant impact on the results. We select a simple prompt that emphasizes that the LLM should generate random addresses and translate them to the languages of each city, so that the requests are in the same language as the streets. The prompt used in English was:

*"Generate a random address in the city of {CITY} ({Country}). Generate a complete address, including the number, and make sure it exists. First, choose the street (or avenue, boulevard, etc.) randomly, trying to ensure that all streets have the same probability, and then the number, trying to ensure that all numbers that exist on that street have the same probability. Return only the complete address, with no additional text. Output only the complete address. Do not include any explanation, reasoning, or additional information. Respond with the address text only."*

Finally, to make the experiment statistically significant and obtain a substantial number of responses, 1,000 repetitions of the same prompt in different Application Programming Interface (API) calls were made for each model and city[3].

**Streets are not Random**

The results of the percentage of times that a given street appears in each model responses are shown in Figure 1 for the top 5 streets. The first observation is that, despite the instructions given in the prompt, the choices are clearly not random. For instance, Madrid has approximately 10,000 streets; in a truly random selection of 1,000 streets, only a few would appear more than once. However, in our results, the most frequent street appears over 200 times, and in one case, more than 600. The same applies for the other three cities, LLMs are clearly concentrating their responses in a few streets and are not capable of sampling uniformly over all streets.

---

[3] All the results are available at https://github.com/aMa2210/LLM_Addresses/tree/main

It is also interesting to see if the streets selected by the LLMs correspond to the most important or iconic streets. This is harder to do as there is some degree of subjectivity in the evaluation. To use a common criterion, we asked the Web version of ChatGPT to list the ten most important streets in the four cities and the results are summarized in Table 2. Then for each of those streets we count the number of times they appear in the responses of the models and visualize the results as a heatmap in Figure 2. For Madrid, and Rome the streets "Calle de Alcala" and "Via del Corso", respectively, have the largest appearances with a large difference from the rest. In case of Paris, Rue de Rivoli is the one with the highest frequency, although with less difference over the rest. Finally, in the case of Amsterdam, there is no street with a high appearance count among the most iconic streets according to ChatGPT. To better visualize the selected streets by the models, those with the top ten counts are shown in a similar heatmap in Figure 3. Among the top ten streets shown in Figure 3, only 27.5% also appear in the list of iconic streets presented in Table 2. Additionally, for Amsterdam, none of the three top streets in Figure 3 appear in Table 2, while for Paris and Madrid this occurs for two and for Rome of one of the top three streets. This shows that the model responses do not always reflect the most iconic streets. This suggests that importance or popularity is not the only factor that influences LLM responses, hinting at the presence of more complex underlying biases [5]. The performance also is generally worse for Amsterdam which may be due to Dutch having a smaller presence in the training datasets of most LLMs.

A possible reason for the biases is that when asking the LLM for an address, the model tends to prefer naming streets over avenues, boulevards, squares, etc. This causes the first output token to be "street" ("calle", "rue", "vie") which already influences the final response. This may result in streets like Gran Vía (the most iconic one in Madrid) not appearing among the top ten results returned by the LLM but other iconic streets like Calle de Alcalá or Calle de Serrano appear with very high frequency. To validate our approach, we computed the log probabilities of the first token generated by the GPT-4.1 model. We found that the token "C" has a probability of 99.87%, whereas the token "Gran" has a probability of less than 0.0005%. This aligns with the model's tendency to respond with the most popular street names, but only for those which names begin with "Calle".

We conducted a manual check for the top streets in Madrid and found that some models are prone to hallucinations. For example, out of the top five streets generated by Qwen3-8B, four of them: "Calle de las Letras", "Calle de las Flores", "Calle de los Ángeles" and "Calle de los Álamos" do not appear in the official list of streets in Madrid [6]. In some cases, they appear as "Plaza" or "Ronda" but not as "Calle" and are not particularly important nor remarkable. Therefore, LLMs may also generate invalid addresses which may create issues for some applications.

In summary, models are unable to randomly pick streets, and their responses are not only influenced by the importance of the street, and some models generate invalid addresses. The results also change significantly when a different prompt is used, although the same general trends are observed. The results and figures when using the same prompt but in English they are available in the repository to illustrate the impact of the prompt.

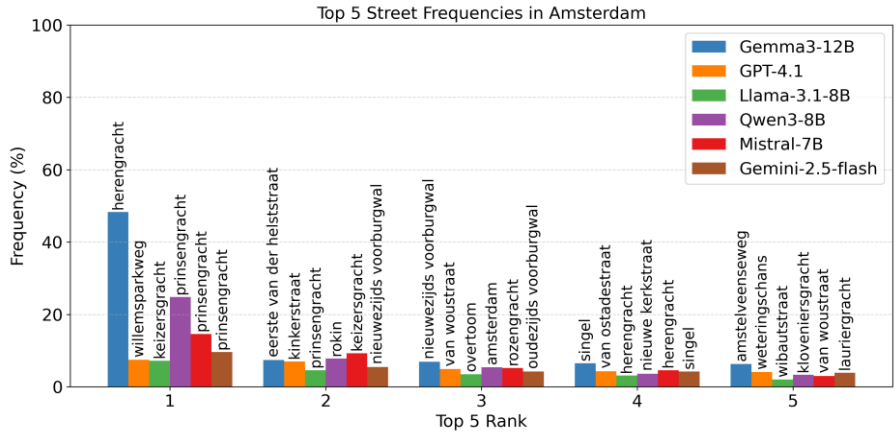
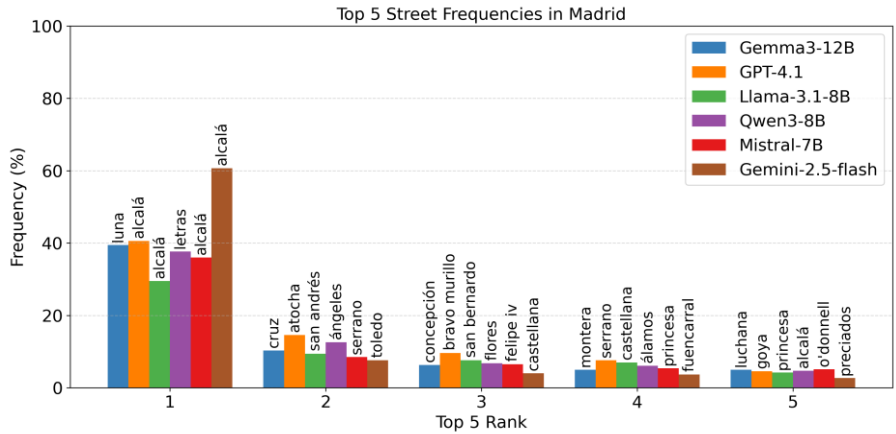
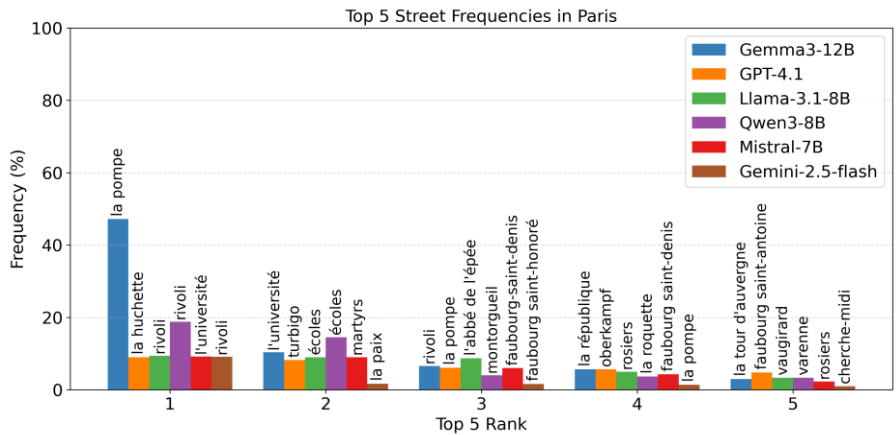
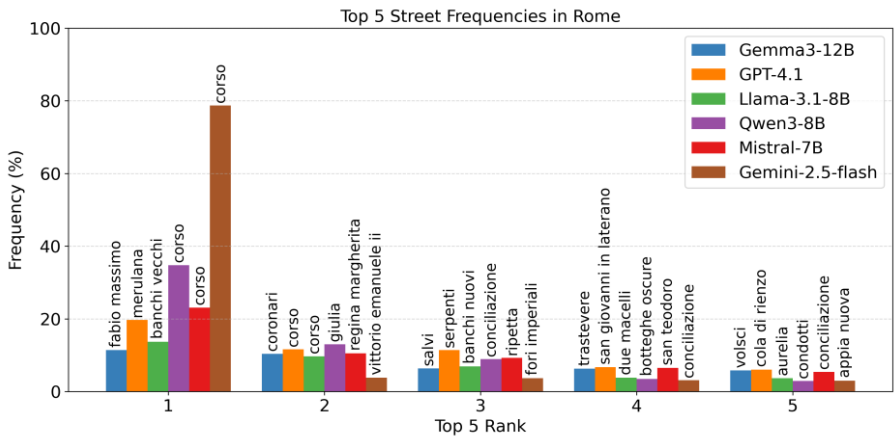

*Figure 1. Top 5 street responses for the models on the four cities*

*Table 2. Most iconic streets according to ChatGPT.*

| Amsterdam | Madrid | Paris | Rome |
|---|---|---|---|
| Kalverstraat | Gran Vía | Avenue des Champs-Élysées | Via del Corso |
| De Negen Straatjes | Paseo de la Castellana | Rue de Rivoli | Via della Conciliazione |
| P.C. Hooftstraat | Calle de Alcalá | Avenue Montaigne | Via dei Fori Imperiali |
| Leidsestraat | Calle de Preciados | Rue du Faubourg Saint-Honoré | Via Condotti |
| Haarlemmerstraat / Haarlemmerdijk | Calle de Serrano | Boulevard Saint-Germain | Via Vittorio Veneto |
| Rokin | Calle Mayor | Avenue de l'Opéra | Via Giulia |
| Damrak | Calle de Atocha | Rue Montorgueil | Via dei Coronari |
| Nieuwendijk | Calle de las Huertas | Boulevard Haussmann | Via del Pantheon |
| Utrechtsestraat | Paseo del Prado | Rue Mouffetard | Via Nazionale |
| Brouwersgracht | Calle de Fuencarral | Rue de l'Abreuvoir | Via Cola di Rienzo |

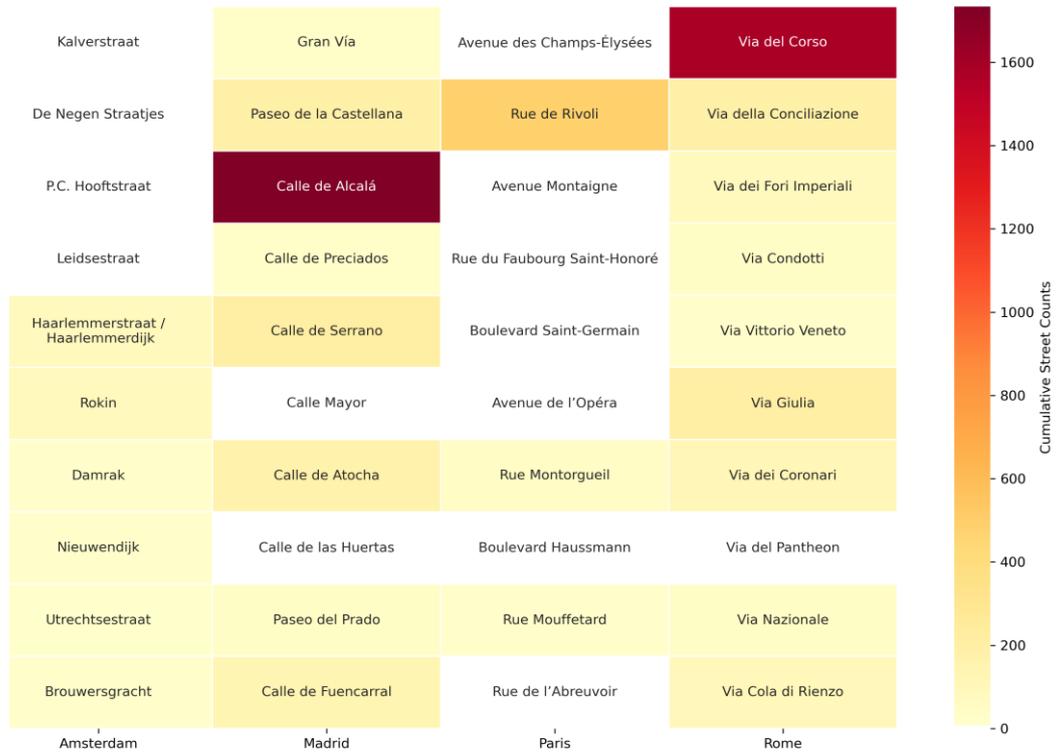

*Figure 2. Number of times that the most iconic streets appear in the LLM responses*

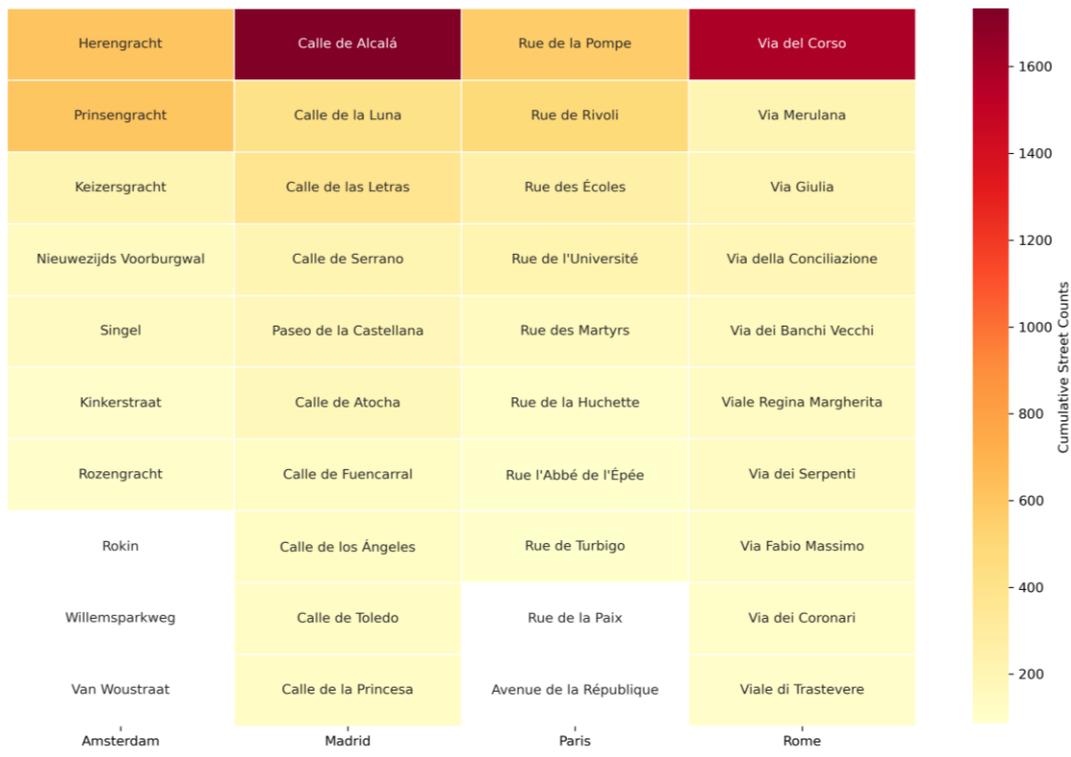

*Figure 3. Number of times of the 10 streets with higher counts in the LLM responses*

**Numbers do better, in some cases**

Unlike street names, address numbers are generally free from factors such as popularity, semantic bias, or cultural associations, and are thus expected to be generated more uniformly. Our findings summarized in Figure 4 largely confirm this conclusion: in most cases, models produced more evenly distributed address numbers than street names. The variance was smaller for numbers than for streets in 15 out of the 24 combinations for parameters city and model. The cases on which variance is larger for numbers than for streets are concentrated in Paris (6 out of 6) and Qwen3 (3 out of 4) but still for most cities and models, numbers are more evenly distributed than streets.

However, it is worth noting that Gemma3 and Qwen3 exhibited a disproportionately high frequency of selecting the numbers "14" and "12" respectively when generating addresses in Paris, and "17" and "45" in Madrid. Beyond the general challenge that LLMs face in producing truly random numbers, this phenomenon may also be attributed to the specific address format used in Paris. For example, a typical address appears as "17 Rue de Charonne, 75011 Paris, France" where the address number comes at the very beginning. This means that the model selects the number before selecting the street which limits the model's ability to introduce randomness incrementally through its autoregressive generation process. Instead, it must sample the number at the beginning of the process from a fixed probability distribution, potentially amplifying the underlying biases.

Interestingly, the models have also shown a preference for specific numbers, such as "123", "12", and "14". These favored numbers were not linked to any specific well-known addresses; rather, they appeared broadly across various generated street names, suggesting a more general bias within the models.

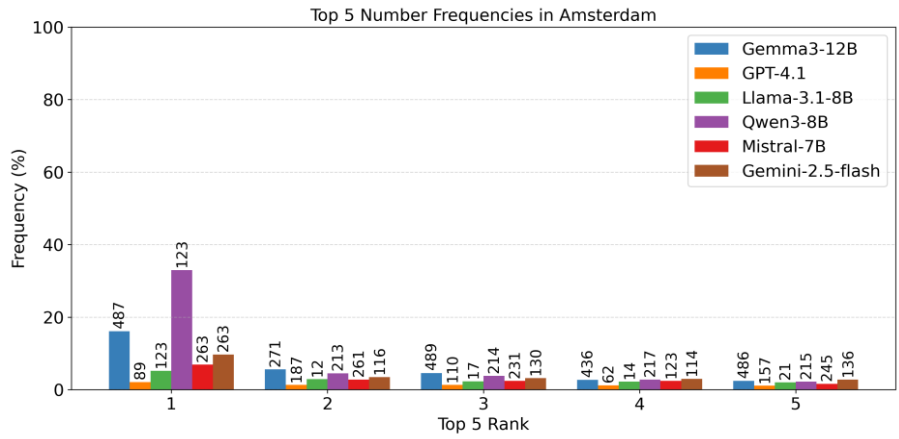
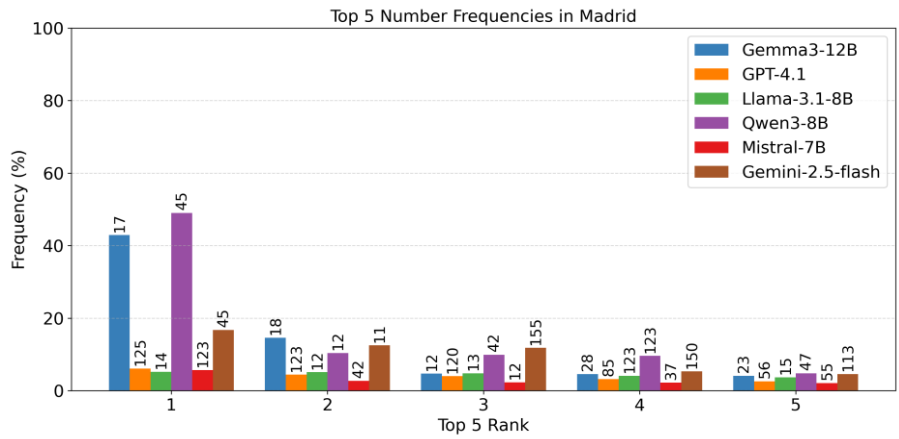
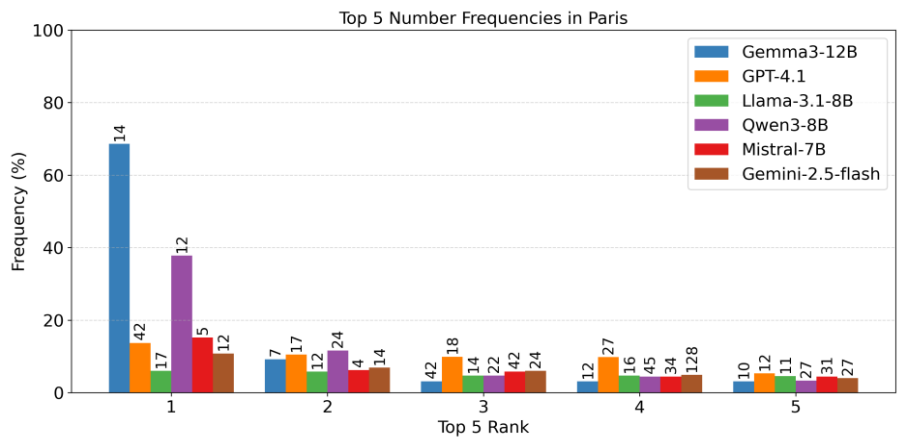
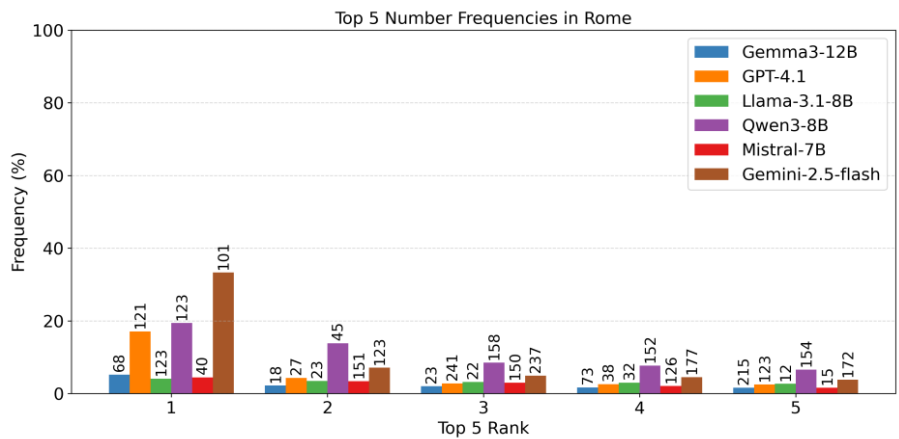

*Figure 4. Top 5 numbers responses for the models on the four cities*

**A Map is worth a thousand words**

To combine streets and numbers and visualize the results, the LLM responses have been plotted on a map of the city for each model as heatmaps using OpenStreetMap[4]. As an example, the maps for Gemini and Qwen3 in Madrid are shown in Figure 5. Gemini produces a more even distribution of the addresses across the city while the responses for Qwen3 are concentrated on two points: Calle de Alcalá 123 and Paseo de la Castellana 123 which shows the bias in the selection of the numbers. The limited number of points in Qwen3 is partly because many of the addresses generated by Qwen3 are invalid and thus OpenStreetMap is unable to plot them. Visualization on a map can help to better understand the biases and trends of LLM when generating addresses and also to compare the responses of different LLMs. The plots of all combinations of cities and models are available in the repository for further analysis.

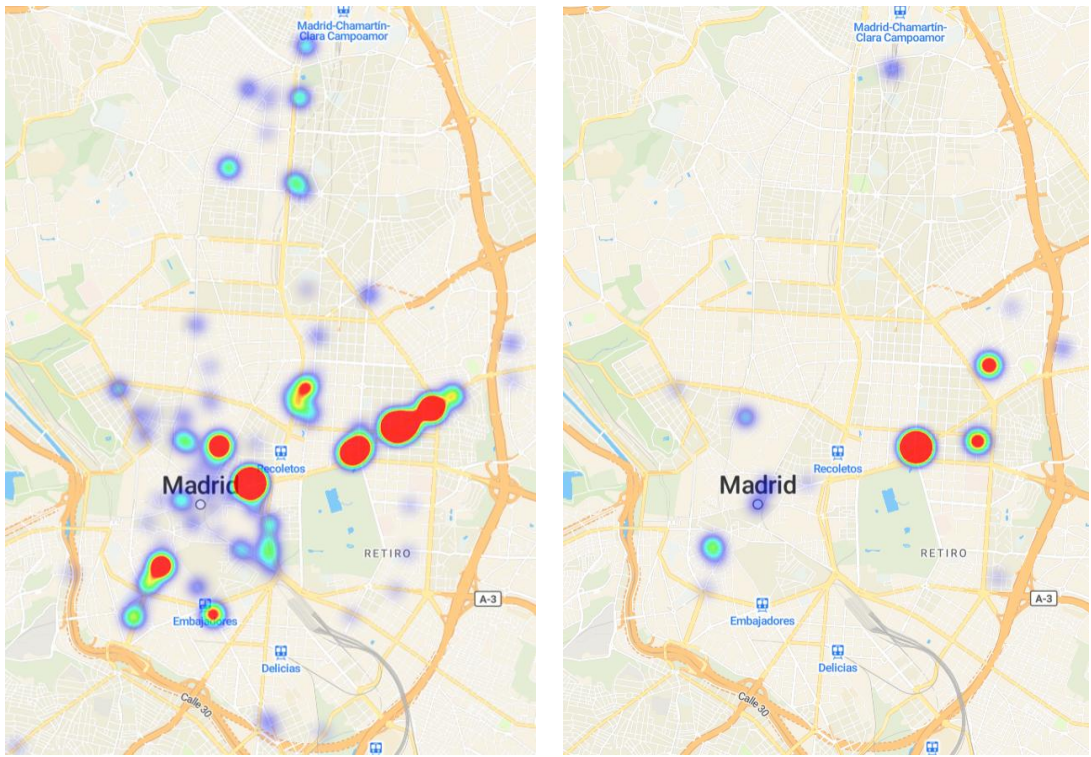

*Figure 5. Madrid heatmaps for Gemini (left) and Qwen3 (right)*

**Conclusion**

In this column we have shown that LLMs are unable to generate random addresses: their responses are concentrated on a few streets and the numbers have also significant biases. This illustrates another limitation of LLMs on a simple task. The next step will be to check if the models can learn to produce more random like addresses if they are fine-tuned with examples [2] and if so, whether the improvement generalizes to other cities. The study of the biases in the results is also of interest to better understand how the training data influences model behavior.

---

[4] https://www.openstreetmap.org/

**Acknowledgements**

This work was partially supported by the Spanish Agencia Estatal de Investigación under Grants FUN4DATE (PID2022-136684OB-C22) and SMARTY (PCI2024-153434), by TUCAN6-CM (TEC-2024/COM460) funded by CM (ORDEN 5696/2024) and SMARTY funded by the European Commission through the Chips Act Joint Undertaking project SMARTY (Grant 101140087). The access to OpenAI models was provided by their Researcher Access Program.

TAIRAN FU is a PhD student at Politecnico di Milano, 20156 Milan, Italy. Contact him at tairan.fu@polimi.it

DAVID NAZARENO-CAMPO is a senior technical expert at FIWARE Foundation and a PhD student at Universidad Politécnica de Madrid, Spain. Contact him at dncampo@gmail.com

JAVIER CORONADO-BLAZQUEZ is a senior data scientist at Telefónica Tech, AI & Data Unit, 28050 Madrid, Spain. Contact him at j.coronado.blazquez@gmail.com

JAVIER CONDE is an assistant professor at Universidad Politécnica de Madrid and a Researcher at the Information Processing and Telecommunications Center, 28040 Madrid, Spain. Contact him at javier.conde.diaz@upm.es

PEDRO REVIRIEGO is a professor at Universidad Politécnica de Madrid, 28040 Madrid and a Researcher at the Information Processing and Telecommunications Center, Spain. Contact him at pedro.reviriego@upm.es

FABRIZIO LOMBARDI is a professor at Northeastern University, Boston, MA 02115 USA. Contact him at F.Lombardi@northeastern.edu